\documentclass[letterpaper, 10 pt, conference]{ieeeconf}

\IEEEoverridecommandlockouts                              
\newcommand\vertbar{\kern1pt\rule[-\dp\strutbox]{.8pt}{\baselineskip}\kern1pt}
\usepackage{graphicx}
\usepackage{caption}
\usepackage{subcaption}
\usepackage{tikz}
\usepackage{booktabs} 
\usepackage{multirow}
\usepackage{makecell}

\usepackage{amsmath}
\usepackage{amssymb} 
\usepackage{hyperref}
\usepackage{wrapfig}
\usepackage{float}
\usepackage{physics,amsmath}
\usepackage{siunitx}

\title{\LARGE \bf
Expressing and Inferring Action Carefulness \\ in Human-to-Robot Handovers}

\author{
    Linda Lastrico$^{*,1,2}$
    Nuno Ferreira Duarte$^{*,3}$,
    Alessandro Carf\'{i}$^{2}$,\\
    Francesco Rea$^{1}$, 
    Alessandra Sciutti$^{1}$, Fulvio Mastrogiovanni$^{2}$,
    Jos\'e Santos-Victor$^{3}$
\thanks{ Corresponding author: {\tt\small{linda.lastrico@iit.it}}}
\thanks{$*$ Equal contribution}
\thanks{$^{1}$Cognitive Architecture for Collaborative Technologies Unit (CONTACT), Italian Institute of Technology, Genoa, Italy }%
\thanks{$^{2}$Department of Informatics, Bioengineering, Robotics, and Systems Engineering (DIBRIS), University of Genoa, Genoa, Italy}%
\thanks{$^{3}$Vislab, Institute for Systems and Robotics (ISR\vertbar Lisboa), Instituto Superior T\'{e}cnico, Universidade de Lisboa, Portugal}%
\thanks{The paper is partially supported by FCT Project UID/50009/2020, the Lisbon ELLIS unit, LUMLIS, and the Research Infrastructure RBCOG-Lab; by the European Commission within the Horizon 2020 research and innovation program, under grant agreement No 870142, project APRIL (multipurpose robotics for mAniPulation of defoRmable materIaLs in manufacturing processes); by the Italian Ministry of University and Research (MUR) through the RAISE project under grant agreement No. D33C22000970006. AS is supported by a Starting Grant from the European Research Council (ERC) under the European Union’s Horizon 2020 research and innovation programme. G.A. No 804388, wHiSPER.
}
}

\begin{document}

\maketitle
\thispagestyle{empty}
\pagestyle{empty}

\begin{abstract} 

Implicit communication plays such a crucial role during social exchanges, that it must be considered for a good experience in human-robot interaction. 
This work addresses implicit communication associated with the detection of physical properties, transport, and manipulation of objects. 
We propose an ecological approach to infer object characteristics from subtle modulations of the natural kinematics occurring during human object manipulation. Similarly, we take inspiration from human strategies to shape  robot movements to be communicative of the object properties while pursuing the action goals. 
In a realistic HRI scenario, participants handed over cups - filled with water or empty - to a robotic manipulator that sorted them. We implemented an online classifier able to differentiate careful/not careful human movements, associated with the cups' content. 
We compared our proposed ``expressive'' controller, which modulates the movements according to the cup filling, against a neutral motion controller. Results show that human kinematics is adjusted during the task, as a function of the cup content, even in reach-to-grasp motion. Moreover, the carefulness during the handover of full cups can be reliably inferred online,  well before action completion.
Finally, although questionnaires did not reveal explicit preferences from participants, the expressive robot condition improved task efficiency.
\end{abstract}

\renewcommand{\theenumi}{\roman{enumi}}%

\section{Introduction}

When interacting with another agent, coordination emerges from the exchange, often independent from will or consciousness, of implicit signals \cite{Knoblich2011jointaction}. This exchange  gives rise to a series of mutual synchronization and anticipation phenomena that drastically reduce the need for complex verbal instructions and resulting delays \cite{Bicho2010integrating}; reciprocal understanding elicits coordination \cite{Sciutti2018mutual}. Therefore, implicit communication is particularly relevant in Human-Robot Interaction (HRI) field, and robots should be able to express and interpret non-verbal cues. Within this context, we address the challenges associated with object properties detection, transport, and manipulation.

As for deducing object characteristics, instead of relying on external appearance or complex multi-modal features \cite{pang2021towards,Modas2021filling,Iashin2021audiovisual,Dimiccoli2022objmaterial,apicella2022container}, we propose an ecological approach that learns from human strategies \cite{duarte_human_2020,lastrico_careful_2021}. The idea is to interpret those kinematics modulations that occur when interacting with items, which have been observed, for instance, in the manipulation of objects of different weights \cite{Campanella2011observegrasp, Flanagan2007weight}.

Coming to the generation of expressive movements with robots, communication is seen as an additional layer that modulates the action without changing its goal \cite{Knepper2017implicitJoint,Dragan2015collaboration, pezzulo_human_2013}. Such a layer needs to be carefully designed: if the robot's motion contradicts biological traits in joint action, such as by adopting an incorrect velocity profile or timing, the human ability to anticipate will be compromised, resulting in a slow and inefficient interaction \cite{Curioni2019jointaction}. Timing modulation plays an important role even in a simple robotic action such as carrying a cup, inducing observers to interpret it as more or less confident, natural, animated, or involving heavier objects \cite{Zhou2017timing}. 

    \begin{figure}[t]
      \centering
      \includegraphics[width = 0.4\textwidth]{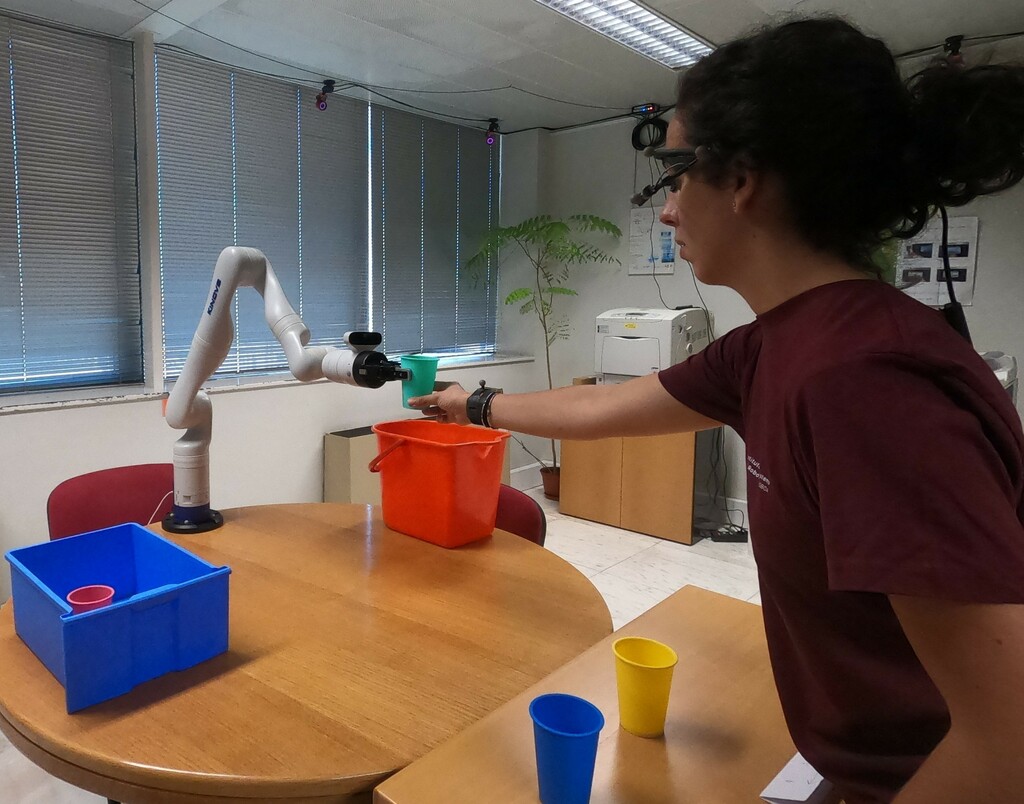}
      \caption{A human handover of a cup with water to a robotic arm. The robot pours the water into the (orange) bucket and then places the emptied cup into the (blue) drawer. Human is wearing an eye-tracking device, infrared motion tracking markers, and an IMU sensor on the wrist} 
      \label{fig:first}
	\end{figure}
\begin{figure*}[t]
\centering
\begin{subfigure}[]{.3\textwidth}
    \centering
    \includegraphics[width=\textwidth]{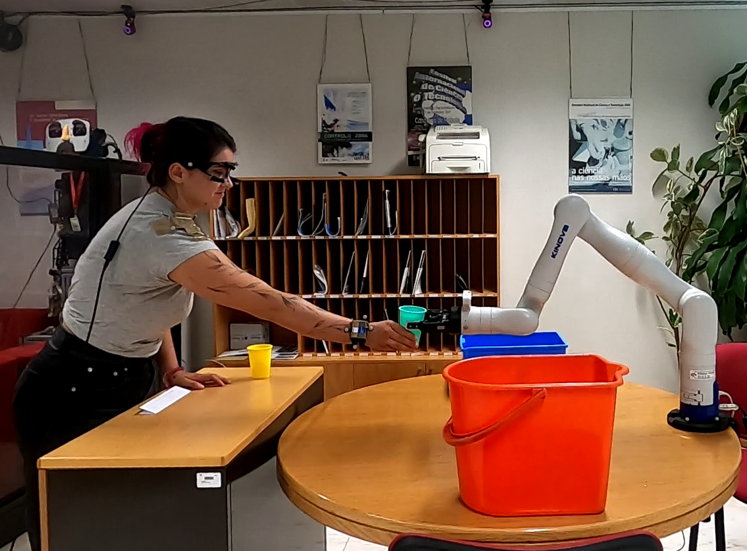}
    \caption{Handover}
    \label{fig:handover}
\end{subfigure}
\begin{subfigure}[]{0.3\textwidth}
    \centering
    \includegraphics[width=\textwidth]{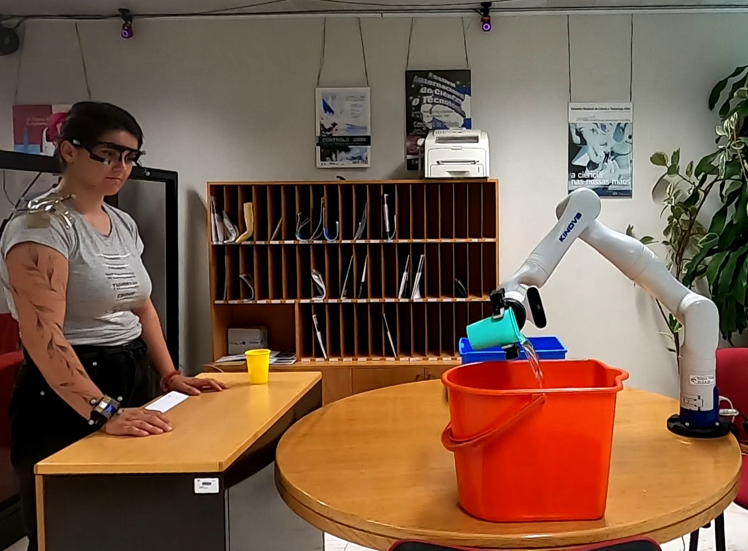}
    \caption{Pouring}
    \label{fig:pour}
\end{subfigure}
\begin{subfigure}[]{0.3\textwidth}
    \centering
    \includegraphics[width=\textwidth]{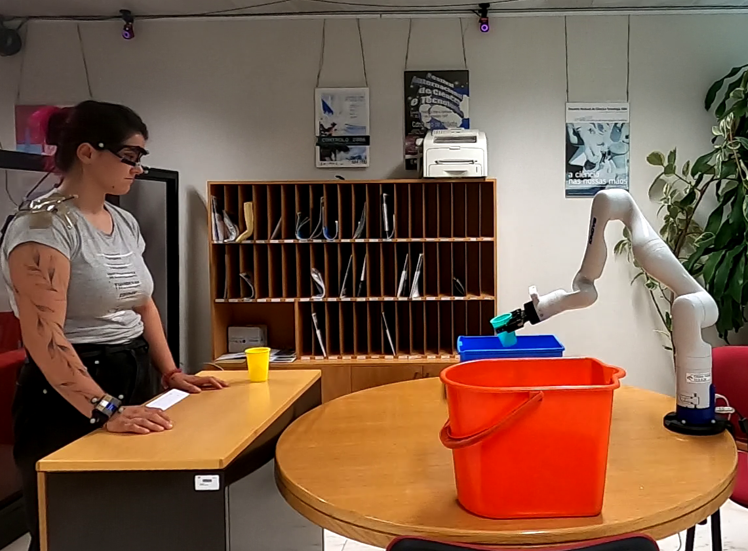}
    \caption{Dropping}
    \label{fig:drop}
\end{subfigure}

\caption{Sequence of actions by the human and robot. From left to right: human-to-robot handover of a full cup (\ref{fig:handover}); robot pouring the water content into the bucket (\ref{fig:pour}); robot placing the emptied cup in the box (\ref{fig:drop}). When manipulating an empty cup, after the handover (\ref{fig:handover}), the robot directly drops the cup in the blue container (\ref{fig:drop})}
\label{fig:HRI}
\end{figure*}

With this work, we propose a novel and realistic interaction between humans and robots where the parts have to collaborate in handling cups with different content: empty or filled with water almost to the brim (see Figure \ref{fig:first} for reference). The robot receives the cups from the human and sorts them, adopting a neutral motion controller or an expressive one that generates movements designed according to the cup content, depending on the experimental condition. We evaluate human kinematics before and during the object manipulation, to observe how the movement is modulated in the context of interaction with a robot and expand previous knowledge on careful handling \cite{duarte_human_2020,lastrico_careful_2021}. We validate the possibility of detecting the carefulness in human-to-robot handovers online, relying on kinematic cues that characterize human movements \cite{duarte_role_2022}. We deploy a human-inspired motion controller based on a generative model, which proved effective for implicit communication and even solicited motor contagion \cite{lastrico_if_2022}, in a scenario where the robot needs to handle full and empty cups appropriately. Finally, we compare the neutral and expressive robot conditions in terms of task efficiency and participants' preferences collected through questionnaires. To summarize, we set the following research objectives: 
\begin{enumerate}
    \item study human kinematics associated with object manipulation; 
    \item validate an online carefulness classifier based on the human motion;
    \item implement an expressive motion controller in a human-to-robot handover scenario; 
    \item confront neutral and expressive conditions.
\end{enumerate}

The dataset collected during the experiments has been made publicly accessible. It consists of synchronized data from marker-based motion capture, an inertial sensor on the wrist, an eye-tracker system, and annotated labels that define the main phases in each trial. It also includes the footage of the experiments from an external camera.

\section{Human-Robot Experiment}

This section describes the human-robot interaction experiments, details the sensors used for the dataset acquisition, and presents administered questionnaires. 

\subsection{Scenario} 

Participants stand in front of a table with four identical plastic cups placed in a row, equidistant from each other. These cups differ in content, being two empty and two filled with water almost to the brim, constituting two types of objects to be handover: empty or full. Participants faced a Kinova Gen3 robot fixed to a table with two distinct recipients at the robot side. As seen in Figure \ref{fig:first}, on the left side of the robot, there is an orange bucket meant to contain water, while the blue drawer on the right stores the empty (or emptied) cups. 

The experiment is presented as a collaborative task, where the human should help the robot clean the table by handing over the cups, from the rightmost cup to the leftmost, one at a time. The robot receives the cup (see Figure \ref{fig:handover}); in case the cup contains water, it pours the content into the orange bucket (Figure \ref{fig:pour}), and finally, it places the empty cup in the blue drawer (see Figure \ref{fig:drop}). 

We adopted a \textit{within-subject} study design where participants are exposed, in a randomized order, to two conditions associated with the controller used by the robot to complete the task: a neutral motion and an expressive motion. More details on the motion controllers will be provided in Section \ref{sec:technical}. In each condition, participants completed 12 handovers to the robot, divided into three blocks, with the experimenter resetting the setup and the four cups on the table at the end of each block. The sequence of empty and full cups to be handed over was balanced but changed in every block. The naturalness of the human-to-robot handover was made possible by constantly tracking the 3D wrist pose with a motion capture system, without requiring a pre-determined and fixed location. The robot's gripper position was computed using the forward kinematics, and a threshold was set to grasp the cup when the distance between the gripper and the participant's wrist was below a certain value.

\subsection{Sensors and Data Description}

Movements' kinematics and visual information were recorded throughout the experiment. Pupil Labs head-mounted glasses \cite{kassner_pupil_2014} were used to track eyes' movements providing 2D gaze fixation on a POV perspective and information on pupil dilation. For more information on the Pupil Labs data acquisition, consult the repository\footnote{Pupil-Labs software repository:\\ \href{https://docs.pupil-labs.com/core/software/pupil-player/\#raw-data-exporter}{https://docs.pupil-labs.com/core/software/pupil-player}}. An OptiTrack motion capture (MoCap) system, consisting of 12 infrared cameras around the room, tracked the position of head, shoulder, and wrist, through reflective rigid bodies suitably designed. Additionally, an Inertial Measurement Unit (IMU), LPMS-B model\footnote{IMU sensor datasheet: \href{https://www.lp-research.com/wp-content/uploads/2013/06/LpmsBUsersGuide1.2.7.pdf}{https://www.lp-research.com/wp-content/uploads/2013/06/LpmsBUsersGuide1.2.7.pdf}}, was placed on the participants' wrists. The acquisition with the three sensors was synchronized through ROS, providing a synchronized timestamp. Moreover, the main events in each trial were manually annotated during the experiment, providing labels for human grasping, handover, robot pouring, and object release. Finally, an external RGB camera recorded the experiments from the viewpoint pictured in Figure \ref{fig:HRI}. Table \ref{tab:dataset} presents the sampling frequency for each sensor and the corresponding total amount of data\footnote{The human-to-robot \textit{handover} actions dataset is publicly available on the institution's  \href{https://vislab.isr.tecnico.ulisboa.pt/datasets_and_resources/\#hcups_water}{website.}}.

Our study involved 15 right-handed participants ($8$ females, $7$ males, $26.6\pm6.2$ years old) who provided written informed consent. They were all naive regarding the purpose of the experiments and not directly involved in our research. The self-reported level of knowledge in robotics was: $40.0\%$ professional or advanced, $33.3\%$ average, and $26.7\%$ little or none. 
360 actions were recorded and performed successfully without dropping the cup or spilling the content. Due to ambiguities or missing data from the MoCap system, in the kinematic analyses of the current study, we considered a total of 310 trials, $76$ and $82$ handovers for empty cups, respectively, in neutral and expressive conditions, and $70$ and $82$ trials, involving full cups for neutral and communicative, respectively. 

\begin{table}[t]
    \centering
    \caption{Sensor specifications and size of recordings.}
    \label{tab:dataset}
    \begin{tabular}{lccc}
          \hline
            Sensor & Type of data & Frequency (Hz) & Total Size \\ \hline
            OptiTrack & Motion Tracking & 120 & $\sim$1.1 GB\\
            Pupil Labs & Eye Tracking & \{30, 120\} & $\sim$100 GB \\
            LPMS-B & IMU & 400 & $\sim$320 MB\\
            GoPro & Video output & 60@1080p & $\sim$40 GB \\
          \hline
          \multicolumn{4}{c}{\footnotesize Note: The PupilLabs streams 30 Hz@720p for the gaze fixation } \\
          \multicolumn{4}{c}{\footnotesize and 120 Hz@320p for the pupil detection system of each eye.}
    \end{tabular}

\end{table}

\subsection{Questionnaires}
Together with a quantitative kinematic description of the human motion, we were interested in investigating if and how the controller conditions would affect how participants \textit{explicitly} perceived the task. For this reason, we conducted a pre-questionnaire to understand the general perception and propensity to robotics; we then administered the same set of scales after each of the two robot controller conditions and finally conducted a post-questionnaire to assess the global perception of the experimental conditions. All the items were based on a 5-point Likert scale, except for the post-questionnaire. Indeed, at the end of the experiment, we asked participants to think about the two interactions experiences they had with the robot and express their preferences concerning a list of 9 items, such as ``In which one of the two parts you were more comfortable in interacting with the robot?'' or ``Which one of the two parts of the interaction you enjoyed the most?''. We also included a speed-bump item in the questionnaires after each interaction to check the participants' level of attention during the completion. See Section \ref{res:questionnaires} and Table \ref{tab:questionnaires} for further details on the scales.
We used Jamovi\footnote{\url{https://www.jamovi.org/}} to analyze the data collected in the survey by using Wilcoxon rank-sum tests, correlation analysis, and binomial proportion tests. 

\section{Human-Robot Communication} \label{sec:technical} 
We designed two robot conditions, neutral and expressive, to compare the efficacy and the effect on the interaction of adding a communicative layer to the generated actions. Moreover, we assess the detection of carefulness in the human manipulation of cups.

\subsection{Robot Motion Controller}\label{sec:robot_controller}
We use a 7 Degrees of Freedom (DOF) Kinova Gen3 robotic manipulator with the Robotiq 85 two-finger gripper\footnote{Official website of the gripper:\\ \href{https://robotiq.com/products/2f85-140-adaptive-robot-gripper}{https://robotiq.com/products/2f85-140-adaptive-robot-gripper}} to accomplish the challenging task of handling cups that may be filled with water. Our architecture is implemented in ROS with the package kortex$\_$ros\footnote{Official repository of the Kinova Gen3 ROS package: \href{https://github.com/Kinovarobotics/ros\_kortex}{https://github.com/Kinovarobotics/ros\_kortex}}, which provides a velocity controller in Cartesian space, moving the end-effector at 40 Hz frequency for linear (m/s) and angular (rad/s) velocities. 

Our primary interest is to test and deploy a controller that mimics what happens naturally in human manipulations, being communicative of the object properties while remaining task-oriented. Its expressive ability has already been tested in a previous study \cite{lastrico_if_2022}, where participants could grasp the implicit message conveyed by the robot's gestures and even showed the effects of motor contagion, although the robot was not humanoid. In this study, participants act instead as givers and perfectly know the characteristics of the object beforehand. Hence, the impact of the robot's implicit communication is reduced. However, even in a context where the manipulator's role is passive, we hypothesize that human-inspired motions can benefit the interaction, making it more fluid, smooth, and desirable to the human partner.
\subsubsection{Neutral condition (NEU)}
In the neutral condition, the actions were always designed with the same approach, generating a constant velocity profile throughout the trials with a simple proportional controller. 
The value of the adopted velocity was empirically chosen to be suitable for transporting all the cups. The choice of always using the same constant velocity granted the successful transport of all the full cups without any content spilling but was not optimized for the transport of the empty ones, which posed no risk instead.
\subsubsection{Expressive condition (EXP)}
To produce the robot's actions adapted to the properties of the object involved - \textit{careful} if full of water, \textit{not careful} if empty - we modulated the end-effector velocity, using velocity profiles norms generated by the Generative Adversarial Networks (GANs). Such a model, trained on human motions, can synthesize novel and meaningful velocity profiles belonging to the desired class of motion and avoid a mere copy of the human (see \cite{lastrico_if_2022, garello_property-aware_2021, yoon_time-series_2019} for further details on the model, its training and its deployment in robotics). Another advantage of using synthetic velocity profiles is to capture the natural variability of human actions and avoid producing stereotyped motion. 
However, in this specific study, to avoid introducing a not controlled variable that may impact the comparison with the neutral condition, we randomly selected one velocity profile for careful and one for the not careful movement to be replicated by the Kinova robot.
We used a careful attitude to transport the full cup from the handover position to the orange bucket, where the robot emptied the content. The designed not careful velocity profile was used instead to take the empty cups from the handover position or the orange bucket to the blue drawer where the cup is released.

\subsection{Carefulness Detection Controller}

To detect the carefulness of the human manipulation of cups during the transport phase in handovers, we used the carefulness detection controller presented in \cite{duarte_role_2022}. It first learned a model of two types of motion behaviors, careful and not careful, trained on human handovers of cups in two conditions, empty and full of water. The model input $\textnormal{x} \in \mathbb{D} \subset \mathbb{R}^{+}$ is the norm position $||\vec{\textnormal{x}}|| = \sqrt{x^2 + y^2 + z^2}$ and denotes the distance between the human wrist and the handover location. The model is defined as a mixture of Gaussians such that $\dot{\textnormal{x}} = \pmb{\textnormal{f}}(\textnormal{x})$, one for the not careful and another for the careful behavior. The model used in this paper is the one from \cite{duarte_role_2022}, which showed the best accuracy for detecting full cup handovers as careful behaviors and empty ones as not careful behaviors.

The controller reads the human wrist during the human-to-robot handover, and it can classify online the carefulness of the motion by comparing the learned model behaviors and the human wrist position. The classification runs in real-time by taking at each time step $t$ $$X = \frac{\dot{x}^t - \dot{x}^{t-1}}{x^t - x^{t-1}} \hspace{2cm} Y_i = \frac{\vec{v}_{\dot{x}}}{\vec{v}_x} $$ where $X$ is the input wrist data and $Y_i$ are the learned \textit{careful} and \textit{not careful} eigenvector components $\vec{v} = [\vec{v}_{\dot{x}}, \vec{v}_x]^T$ from the Gaussian covariance matrix $\Sigma \vec{v} = \lambda \vec{v}$. The classifier is based on the belief system $B$ \cite{khoramshahi_dynamical_2019} where $B = [b_1, b_2]$ is initialized as $[50\%, 50\%]$ and it computes the error $e =X - \sum_{i=1}^{2} b^t_i Y_i$. $B$ is updated as $b_i^{t+1} \leftarrow b_i^t + \dot{b}^t_i \Delta t$ given $\dot{b}^t_i = \epsilon (e^T + (b^t_i - 0.5)Y_i^2)$ and $\epsilon \in \mathbb{R}^+$. During the cup transportation, the classification will reach either $b_1 =100\%$ (not careful) or $b_2 =100\%$ (careful). However, the robot does not use this information to produce appropriate movements during the interaction. Indeed, we want to evaluate independently from the classification outcomes the generative expressive motion controller against the non-biological neutral one. This is to prevent unpredictable false negative classifications that may result in catastrophic water spills compromising the interaction.

\section{Results and Discussion}
In this section, we will present our findings relative to the analysis of human kinematics during the task, the performances of the online carefulness classifier in detecting the human attitude in both the reaching and transport movements, the results from the questionnaires, and the metrics assessing the task efficiency.

\subsection{Human motion}\label{res:human}
\begin{figure}[t]
     \centering
     \includegraphics[width=\columnwidth]{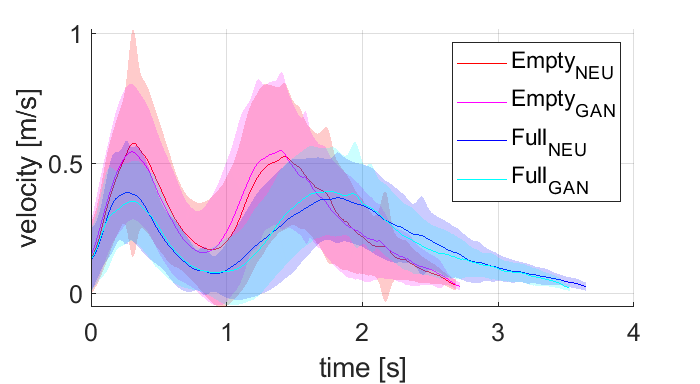}
     \caption{Hand mean velocities and standard deviation, in transparency, associated with the reaching (first peak) and transportation phase (second peak) of full and empty cups, in shades of blue and red, respectively. The robot controller used to transport the cup at a later time did not influence the human behavior as a giver}
    \label{fig:human_velocities}
\end{figure}

Figure \ref{fig:human_velocities} represents participants' hand velocity when moving towards the cup to grasp it (first velocity peak) and then its transportation toward the robot gripper (second peak). This representation results from applying a second-order Butterworth low pass filter with a cut-off frequency of 8 Hz to the hand velocities derived from the 3D trajectories recorded with the MoCap system for each trial. The velocity profiles were resampled to the median duration for each class, and their mean was computed for every time instant, together with the standard deviation. A noticeable difference in the velocity adopted depends on the cup involved: empty cups, associated with not careful motions, elicit quicker motions with higher speed magnitude. Instead, full cups require careful actions to avoid spilling the content, resulting in slower movements with lower accelerations. 
The robot controller does not influence how participants took the cup and handed it over, as could be expected from their role as givers. 
These observations from Figure \ref{fig:human_velocities} were confirmed by running a mixed model with Jamovi GAMLj module\footnote{General analyses for linear models Jamovi module: \href{https://gamlj.github.io/}{https://gamlj.github.io/}} separately on the reaching and the transport actions. We assumed the maximum velocity of participants' movements as the dependent variable, the subjects as cluster variables, and the content of the cup, requiring or not care, the controller type, and their interaction as a factor. 
The effect of condition resulted significant for both the action phases (\textbf{Reaching: }$Full - Empty$, $estimate = -0.202 \unit{\metre\per\second}$, $SE=0.032$, $t=56.6$, $p < 0.001$, \textbf{Transport: }$Full - Empty$, $estimate = -0.276 \unit{\metre\per\second}$, $SE=0.037$, $t=-7.49$, $p < 0.001$). This result shows that the modulation in human action appears not only when directly interacting with the object, when we observe a decrease of the velocity magnitude of $0.276 \unit{\metre\per\second}$ if the cup content is full, but even before touching the object; indeed, participants adapt the motion according to the presence of water, with an estimate of $0.202 \unit{\metre\per\second}$ of velocity reduction when preparing to grasp a full cup. These observations confirm previous results in the analysis of human kinematics associated with the carefulness feature \cite{duarte_human_2020, lastrico_careful_2021}, but interestingly extend the effect of object properties also to the kinematics of reach-to-grasp actions. It is the first time we have observed anticipated motor adaptation in this context; further investigations need to be conducted to understand why this anticipatory effect arose: it may be ascribed to the standing posture of participants or, even more interestingly, to the collaborative setting. Indeed, the intention to communicate and make predictable and readable gestures in a social context is well-described by the signaling theory \cite{pezzulo_human_2013}. For instance, it has been proved that the emerging kinematic pattern in grasping gestures differs between individual and social conditions \cite{Sartori2009intention}.
In any case, the possibility of accessing this implicit information in advance, already in the action preparation, is relevant to plan the activity of the artificial agent accordingly, and its feasibility will be discussed in the following paragraph. 


\subsection{Carefulness Classification}

\begin{figure}[t]
     \centering
     \includegraphics[width=\columnwidth]{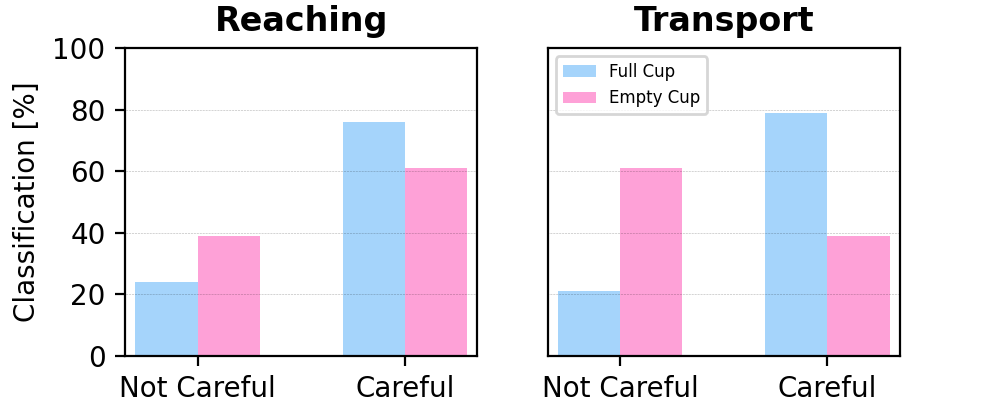}
     \caption{Carefulness classification in reaching and transport motions for both empty and full cup's motions as careful or not careful behaviors. The accuracy is measured as the number of motions with full cups correctly predicted as careful behaviors and the motions with empty cups correctly predicted as not careful behaviors.}
    \label{fig:classify}
\end{figure}

The classification is accomplished using the best model from \cite{duarte_role_2022} with hyperparameter $\epsilon=0.14$. 
It achieved an accuracy of 90\%, 87\%, and 77\% for full cups classified as careful motions, for EPFL \cite{starke_force_2019}, QMUL \cite{qmul}, and IST \cite{duarte2022gaze} datasets, respectively. As for empty cups, they were classified as not careful in 55\%, 50\%, and 52\% for EPFL, QMUL, and IST handovers, respectively. As explained in previous papers \cite{lastrico_careful_2021, duarte_role_2022, lastrico_if_2022} the handover of empty cups does not elicit a cautious behavior as full cups do, i.e. the risk of spilling. As such, the handover of empty cups is freely decided upon the user's handover preference, which may resemble a careful attitude or not and complicate the detection. Figure \ref{fig:classify} show the classification results for the reaching and transport phase. In the transport phase, the accuracy resembles the results obtained previously, with 79\% of full cups classified as careful and 60\% of empty cups as not careful. Considering that this is a human-to-robot handover and the model was trained on human-human handovers, this shows the model's efficacy with new scenarios, participants, and interaction designs. The accuracy in the classification for reaching motions is lower. However, it should be noted that the model was trained on handovers, and it is the physical interaction with the cup that mainly influences the kinematics modulation. Moreover, Figure \ref{fig:human_velocities} shows that the adjustment in the reach-to-grasp action is principally related to the magnitude of the velocity, as mentioned previously, whereas its duration is comparable for empty and full cups, differently to what happens during the transportation phase.

\begin{figure}[t]
     \centering
     \includegraphics[width=\columnwidth]{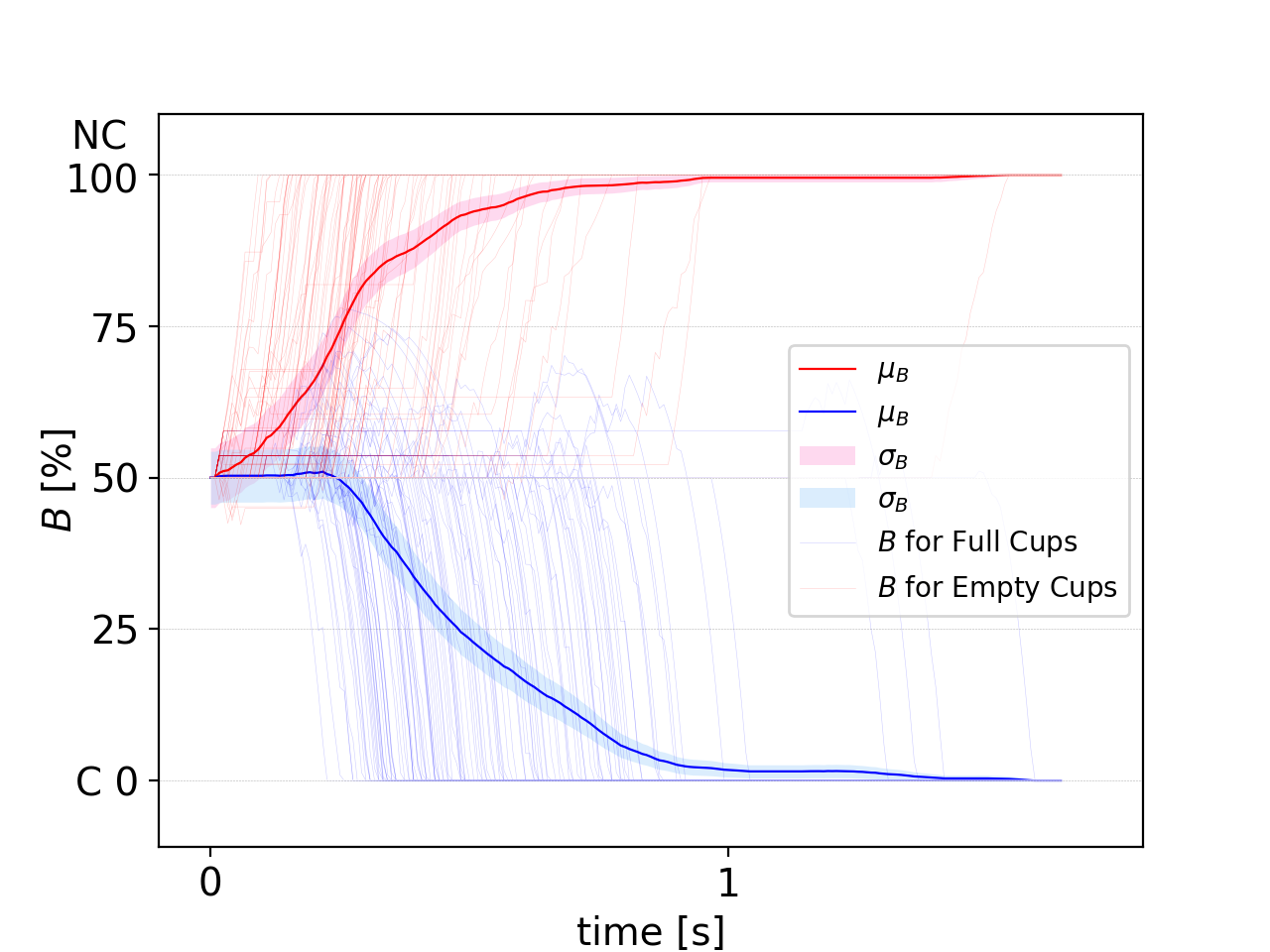}
     \caption{The classifier output $B$ for empty and full cup's motions over time. For each human transport of either cup, the $B$ reaching 0\% or 100\% refers to the classification of C (careful) or NC (not careful) behaviors, respectively. The $\mu$ and $\sigma$ are the mean and the $\sim$84\% confidence interval for the true positive cases of the binomial distribution.}
    \label{fig:reaction_time}
\end{figure}

\begin{table*}[t]
\caption{Questionnaire's scales, N=15}
\centering
\resizebox{\textwidth}{!}{%
    \begin{tabular}{l l l l l l} 
    \toprule
    \textbf{Scale} & \textbf{Item example} & \textbf{Condition} &  \textbf{Cronbach's $\alpha$} & \textbf{Mean$\pm$SD} & \textbf{p}  \\
    
    \midrule     
    Positive Attitudes About Robots (PARS) \cite{bernotat2021keep} &  \makecell[l]{I think the use of robots can have a positive impact on society} & Pre & .83 & $4.50\pm0.57$ & - \\ 
    Negative Attitudes About Robots (NARS) \cite{NARSNomura_2006} & \makecell[l]{I would feel uneasy if I was given a job where I had to use robots} & Pre & .02 & - & - \\ 
    \makecell[l]{Anxiety (Anx.) \cite{anxietyNomura_2006}} & \makecell[l]{I would be anxious about the kind of movements the robot would make} & Pre & .72 & $2.33\pm0.73$ & - \\ 
    \midrule  

\multirow{2}{*}{Anxiety (Anx.)  \cite{anxietyHeerink2010}} &  \multirow{2}{*}{\makecell[l]{I was afraid to make mistakes with the robot}} & NEU & .63 & $1.47\pm0.44$ & \multirow{2}{*}{0.23} \\ 
\multirow{5}{*}{} & & EXP & 
 .64 & $1.57\pm1.50$ &  \\ 
 \cmidrule{3-6}
\multirow{2}{*}{Competence (Comp.) \cite{warmthCompetenceFiske_2007}} &  \multirow{2}{*}{\makecell[l]{I think that the robot was competent}} & NEU & .91 & $3.92\pm0.84$ & \multirow{2}{*}{0.90} \\ 
\multirow{5}{*}{} & & EXP & 
 .88 & $3.87\pm0.83$ &  \\ 
 \cmidrule{3-6}
\multirow{2}{*}{Cognitive Trust in HRI
    (Cogn.) \cite{CognAffBernotat_2017}} &  \multirow{2}{*}{\makecell[l]{I would feel a need to monitor the robot's work}} & NEU & .71 & $3.76\pm0.50$ & \multirow{2}{*}{0.06} \\ 
\multirow{5}{*}{} & & EXP & 
 .82 & $3.90\pm0.56$ &  \\ 
 \cmidrule{3-6}
\multirow{2}{*}{Affective Trust in HRI
    (Affect.) \cite{CognAffBernotat_2017}} &  \multirow{2}{*}{\makecell[l]{This robot would act cooperatively}} & NEU & .51 & - & \multirow{2}{*}{-} \\ 
\multirow{5}{*}{} & & EXP & 
 .79 & $4.09\pm0.64$ &  \\ 
 \cmidrule{3-6}
\multirow{2}{*}{Evaluation of Robot Movements
    (Eval.) \cite{EvalMovBernotat_2021}} &   \multirow{2}{*}{\makecell[l]{The robot's movements looked natural}} & NEU & .79 & $4.08\pm0.39$ & \multirow{2}{*}{0.26} \\ 
\multirow{6}{*}{} & & EXP & 
 .83 & $4.13\pm0.39$ &  \\ 
\bottomrule
\multicolumn{5}{c}{\footnotesize Note: The p value, when reported, refers to a Wilcoxon rank sum test between Neutral and Expressive robot conditions}
    \end{tabular}
}
\label{tab:questionnaires}
\end{table*}

The results of the classification for the object transport phase can be further analyzed in Figure \ref{fig:reaction_time}. The carefulness detection controller ran in real-time, and the belief system $B$ is updated at each time step $t$, providing a decision - careful or not careful - before the handover is completed. Figure \ref{fig:reaction_time} shows that after 1 second (120 time steps) $97\%$ of empty cups are classified, and after 157 time steps ($\sim$1.3 seconds) $97\%$ of full cups are classified. In Figure \ref{fig:human_velocities} the handovers lasted $1.62\pm0.5$ and $2.32\pm0.59$ seconds for empty and full, respectively. This provides the robot with the information of whether the human is careful or not before the cup is in the robot's gripper (1 second beforehand or more for longer durations), potentially allowing it to plan accordingly and in advance its actions. In this study, we assume that the reason for being careful has to do with the cup containing a liquid that may spill, however, this has many other applications, such as moving fragile or dangerous objects which due to the aforementioned properties require a more attentive (careful) handling.

\subsection{Questionnaires}\label{res:questionnaires}

\begin{figure}[t]
     \centering
     \includegraphics[width=\columnwidth]{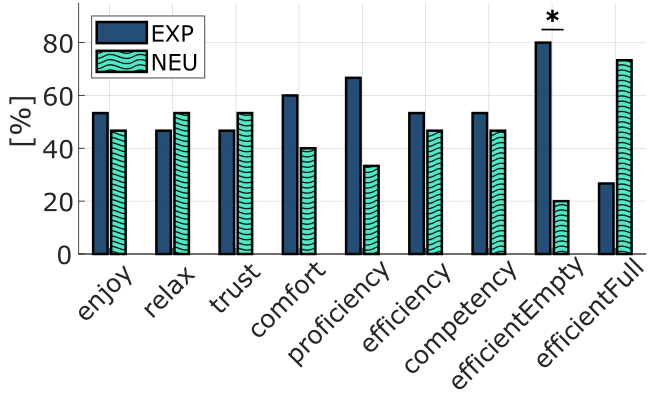}
     \caption{Post-questionnaire preferences between the Neutral and the Expressive GAN-based robot conditions}
    \label{fig:questionnaires}
\end{figure}

The items used in the questionnaires were short versions adapted from validated scales. 
Cronbach's $\alpha$ was used to evaluate the internal consistency and reliability of each of them and is reported in Table \ref{tab:questionnaires}. If the value of $\alpha$ was above $0.60$, unit indices were produced by averaging the responses to the individual items included in each scale. 
The ``NARS'' and ``Affective Trust in HRI'' scales presented a low Cronbach's value ($\alpha = 0.02; 0.51$ respectively); hence they were not considered in the analyses. We also ran Shapiro-Wilk's tests to verify the normality of the samples. When the distribution resulted gaussian, we employed parametric tests; otherwise, we used the non-parametric versions.  
Before starting the experiment, we asked participants to answer a few items to better characterize the population sample. Their mean values are reported at the beginning of Table \ref{tab:questionnaires}. Not surprisingly, we found a positive correlation between the self-reported knowledge of robotics and the PARS scale (Spearman's rho: $0.457, p < 0.05$).
By administering a set of scales after each interaction with the robot, we wanted to assess if the participants perceived the two conditions differently. As previously mentioned, the order of the conditions was balanced and randomized to avoid introducing possible bias; however, as reported in Table \ref{tab:questionnaires}, we found no significant difference. Only the Cognitive Trust in HRI scale reported a slight preference for the expressive condition, but still not statistically meaningful. Considering separately single items and using again a Wilcoxon rank sum test to compare the two conditions, we found two significant differences: ``This robot would act consistently'' ($NEU$: $3.73\pm1.03$, $EXP$: $4.20\pm0.68,\,p<.05$); ``The robot moved too slowly'' ($NEU$: $2.80\pm1.20$, $EXP$: $2.20\pm0.78,\,p<.05$). 
The percentages associated with participants' preferences in the post-questionnaire, after experiencing the two experimental conditions in random order, are reported in Figure \ref{fig:questionnaires}. In this case, the distribution of the answers was binomial, so we ran a two outcomes proportion test. We found a significant difference in the item ``In which one of the two parts was the robot more efficient in transporting the empty cups?'' ($p<.05$). The robot was perceived as more efficient when controlled with the Not Careful GAN's velocity profile. We can also observe a tendency to choose the neutral controller instead when dealing with the full cup. This can be understood by referring to Section \ref{sec:robot_controller}, where we explained that in the design of the neutral condition, we chose a velocity profile intermediate between careful and not that still allowed to complete the task successfully with no spilling; in this sense, the neutral robot was faster when transporting the cup with water. A trend is also noticeable in the Comfort and Proficiency items, where the participants well received the expressive controller.\\
The general questionnaire's outcome does not allow us to conclude about a strong preference of participants between the two conditions. The experiment design required the two agents to collaborate in any case, and the drive to comply and adapt may have overshadowed the differences in robot movements. At the end of the experiment, 12 participants out of 15 reported perceiving a distinction between the two blocks, but only 7 ($46.7\%$) attributed it to a difference in the velocity; the others referred, for instance, to the pouring angle, the gripper strength, or the position of emptying or handover, which did not actually change. However, we can further reason that participants did not consciously recognize or, at least, did not explicitly appreciate the added value of communicative movements. In a previous study, where the robot had the active role of giver, its human-inspired actions accomplished the goal of communicating carefulness associated with object properties and even elicited motor adaptation \cite{lastrico_if_2022}. In this scenario, the participants' role as givers made them fully aware of the object's properties. In this sense, the robot had nothing to communicate except to accomplish the task without failures, which happened in both conditions. Therefore, in such a collaborative task, the robot must first function properly to be accepted, while communicative movement is not necessarily perceived overtly as better.

\subsection{Global metrics}
\begin{figure}[t]
     \centering
     \includegraphics[width=0.75\columnwidth]{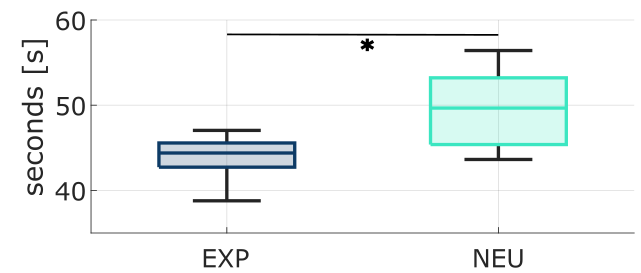}
     \caption{Difference in the block duration considering only the human contribution}
    \label{fig:duration}
\end{figure}

One of our hypotheses when designing a human-inspired robot controller was that it would improve the smoothness and efficiency of the interaction when adopting the appropriate level of carefulness. For this reason, we decided to examine a quantitative metric: the interaction duration over the block of four consecutive cup handovers. In this case, we are interested in a fluency metric that, regardless of the time taken by the robot, which can vary depending on the condition and the cup handed, measures the human efficiency in the task. Hence, we annotated from videos the duration of the robot's action. We considered the net time by subtracting from the total duration of each block the time taken by the robot's movements from each completed handover to the end of the trial, i.e. when it released the empty glass into the blue container. Figure \ref{fig:duration} illustrates the corresponding results, where we find a significant difference in the net duration of the blocks, in seconds, confirmed by a Wilcoxon rank sum Test ($NEU$: $49.5\pm4.35$, $EXP$: $44.4\pm3.06$, $p<.001$). This result is extremely interesting because it proves that, even if not consciously perceived by participants, the expressive modulation of the robot's actions makes the overall interaction smoother and more fluid, with a reduction of 5 seconds on the net duration. The impression we had observing the videos is that when the robot exhibited a human-inspired behavior, the latency time between each trial was reduced. 

\section{Conclusion}
In this study, we presented a realistic scenario where a human and a robot successfully collaborate in handling full and empty cups, with the goal of understanding how to infer human carefulness during handover actions, as well as to generate similarly ``expressive'' robot movements. 
In a dynamic interaction scenario, it may be impractical to predict the contents of every single cup directly from an object's appearance or sound features \cite{pang2021towards,Modas2021filling,Iashin2021audiovisual}.
In our approach, the key to successfully inferring the cup properties relies on the interaction itself. We explore the natural adaptations that occur in human motion when handling cups with different filling levels which we observed, for the first time, not only during the transportation but also in the reach-to-grasp movements. This result, together with the ability of the classifier to infer an action's carefulness by observing just its initial part, allows the robot to prepare the most appropriate motion strategy in advance. We found interesting insights comparing a proportional neutral controller and our human-inspired expressive one. 
%
%
When the robot was not expressive, keeping the same velocity for the whole experiment regardless of the cup content, participants perceived it as less consistent and found its actions generally too slow. 
Compared to an oblivious robot that makes no assumptions about the cup's properties, our approach mimics the human strategy: being careful when needed and speeding up otherwise. Although participants have not acknowledged this difference explicitly, we found a quantitative measure of the interaction efficiency in the experiment duration. Indeed, with the robot showing human-like behavior, humans reduce the time required to complete their tasks, thus improving the fluency and smoothness of the collaborative interaction.

To conclude, this study validates a framework for action perception and generation in HRI. In future works, the complete architecture should be tested in a dyadic interaction where both the human and the robot have an active role in carrying objects. In such a context, the classification output could be directly used to plan adequate robot motions with GANs.



%

\bibliographystyle{ieeetr}
\bibliography{refs, refs_extra}

\end{document}